\documentclass[twoside]{article}
\usepackage[a4paper]{geometry}
\usepackage[latin1]{inputenc} 
\usepackage[T1]{fontenc} 
\usepackage{RR}
\usepackage{hyperref}
\usepackage{graphics} 
\usepackage{graphicx}
\usepackage{epsfig} 
\usepackage{subfigure}
\usepackage{amsmath} 
\usepackage{amssymb}  
\usepackage{latexsym}

\usepackage{bm}             
\usepackage{epstopdf}

\RRNo{8765}

\RRdate{July 2015}

\RRauthor{
K. Avrachenkov\thanks{Corresponding author. K. Avrachenkov is with Inria Sophia Antipolis, 2004 Route des Lucioles, 06902,
Sophia Antipolis, France {\tt\small k.avrachenkov@inria.fr}}%
  \and
P. Chebotarev\thanks{P. Chebotarev is with Trapeznikov Institute of Control Sciences of the Russian Academy of Sciences, 65 Profsoyuznaya Str.,\x{4} Moscow, 117997, Russia}%
  \and
A. Mishenin\thanks{A. Mishenin is with St. Petersburg State University, Faculty of Applied Mathematics and Control Processes,
Peterhof, 198504, Russia}
\thanks[sfn]{This work was partially supported by Campus France,
Alcatel-Lucent Inria Joint Lab, EU Project Congas FP7-ICT-2011-8-317672, and RFBR grant No.\,13-07-00990.}
}
\authorhead{K. Avrachenkov \& P. Chebotarev \& A. Mishenin}

\title{Semi-supervised Learning with Regularized~Laplacian}

\RRtitle{L'Apprentissage Semi-supervis\'e avec Laplacian~R\'egularis\'e}
\RRetitle{Semi-supervised Learning with Regularized~Laplacian}
\titlehead{Semi-supervised Learning with Regularized~Laplacian}

\RRresume{
Nous \'etudions une m\'ethode d'apprentissage semi-supervis\'e, bas\'e sur le graphe de similarit\'e
et Laplacian r\'egularis\'e.
Nous formalisons la m\'ethode comme un probl\`eme d'optimisation convexe et quadratique et
nous \'etablissons ses diverses propri\'et\'es. En particulier, nous montrons que le noyau de la m\'ethode
peut \^etre interpr\'et\'e en termes des marches al\'eatoires en temps discret et continu
et poss\`ede plusieurs propri\'et\'es importantes des mesures de proximit\'e.
Les techniques d'optimisation ainsi que les techniques d'alg\'ebre lin\'eaire peuvent \^etre utilis\'e
pour un calcul efficace des fonctions de classification. Nous d\'emontrons sur des exemples num\'eriques
que la m\'ethode de Laplacian r\'egularis\'e est concurrentiel par rapport aux autres \'etat de l'art
m\'ethodes d'apprentissage semi-supervis\'e.
}
\RRabstract{
We study a semi-supervised learning method based on the similarity graph and Regularized
Laplacian. We give convenient optimization formulation of the Regularized Laplacian method and establish
its various properties. In particular, we show that the kernel of the method
can be interpreted in terms of discrete and continuous time random walks and possesses several important
properties of proximity measures. Both optimization and linear algebra methods can be used for efficient
computation of the classification functions. We demonstrate on numerical examples that the
Regularized Laplacian method is competitive with respect to the other state of the art semi-supervised
learning methods.
}
\RRmotcle{Apprentissage Semi-supervis\'e, Apprentissage bas\'e sur le graphe de similarit\'e,
Laplacian r\'egularis\'e, mesure de proximit\'e, classification des articles Wikipedia}
\RRkeyword{Semi-supervised learning, Graph-based learning, Regularized Laplacian, Proximity measure,
Wikipedia article classification}

\RRprojet{Maestro}

\RCSophia

\newcommand {\1}{\textbf{1}}

\def\x#1{}                      
\def\diag{\operatorname{diag}}                   
\def\F{{\cal F}}
\def\cdc{,\ldots,}
\def\beq{\begin{equation}}                           
\def\eeq{\end{equation}}                             
\def\Ff{\mathsf F}                                   

\begin{document}

\makeRR


\section{Introduction}

Graph-based semi-supervised learning methods have the following three principles at their foundation.
The first principle is to use a few labelled points (points with known classification) together with
the unlabelled data to tune the classifier. In contrast with the supervised machine learning, the
semi-supervised learning creates a synergy between the training data and classification data.
This drastically reduces the size of the training set and hence significantly reduces the cost
of experts' work. The second principal idea of the semi-supervised learning methods is to use
a (weighted) similarity graph. If two data points are connected by an edge, this indicates some
similarity of these points. Then, the weight of the edge, if present, reflects the degree of similarity.
The result of classification is given in the form of classification functions. Each class has its own
classification function defined over all data points. An element of a classification function gives
a degree of relevance to the class for each data point. Then, the third principal idea of the
semi-supervised learning methods is that the classification function should change smoothly
over the similarity graph. Intuitively, nodes of the similarity graph that are closer together
in some sense are more likely to belong to the same class. This idea of classification function
smoothness can naturally be expressed using graph Laplacian or its modification.

The work \cite{ZGL03} seems to be the first work where the graph-based semi-supervised learning
was introduced. The authors of \cite{ZGL03} formulated the semi-supervised learning method as
a constrained optimization problem involving graph Laplacian. Then, in \cite{Zetal04,ZB07} the
authors proposed optimization formulations based on several variations of the graph
Laplacian. In \cite{AGMS12} a unifying optimization framework was proposed which gives as
particular cases the methods of \cite{ZB07} and \cite{Zetal04}. In addition, the general
framework in \cite{AGMS12} gives as a particular case an interesting PageRank based method,
which provides robust classification with respect to the choice of the labelled points \cite{Aetal08,AGS13}.
We would like to note that the local graph partitioning problem \cite{ACL06,C09} can be
related to graph-based semi-supervised learning. An interested reader can find more details
about various semi-supervised learning methods in the surveys and books
\cite{CSZ06,FoussFrancoisseSaerens11,Z05}.

In the present work we study in detail a semi-supervised learning method based on the Regularized
Laplacian. To the best of our knowledge, the idea of using Regularized Laplacian and its kernel
for measuring proximity in graphs and application to mathematical sociology goes back to the works
\cite{CS97,CheSha98}. In \cite{FoussFrancoisseSaerens11} the authors compared experimentally many graph-based
semi-supervised learning methods on several datasets and their conclusion was that the semi-supervised
learning method based on the Regularized Laplacian kernel demonstrates one of the best performances on
nearly all datasets. In \cite{CallutSaerensDupont08} the authors studied a semi-supervised learning
method based on the Normalized Laplacian graph kernel which also shows good performance.
Interestingly, as we show below, if we choose Markovian Laplacian as a weight matrix, several known
semi-supervised learning methods reduce to the Regularized Laplacian method. In this work we formulate
the Regularized Laplacian method as a convex quadratic optimization problem which helps to design easily
parallelizable numerical methods. In fact, the Regularized Laplacian method can be regarded as
a Lagrangian relaxation of the method proposed in \cite{ZGL03}. Of course, this is a more flexible formulation,
since by choosing an appropriate value for the Lagrange multiplier one can always retrieve the method
of \cite{ZGL03} as a particular case. We establish various properties of the Regularized Laplacian method.
In particular, we show that the kernel of the method can be interpreted in terms of discrete and continuous time
random walks and possesses several important properties of proximity measures. Both optimization and linear algebra
methods can be used for efficient computation of the classification functions. We discuss advantages and
disadvantages of various numerical approaches. We demonstrate on numerical examples that the Regularized Laplacian
method is competitive with respect to the other state of the art semi-supervised learning methods.

The paper is organized as follows: In the next section we formally define the Regularized Laplacian method.
In Section~3 we discuss several related graph-based semi-supervised methods and graph kernels.
In Section~4 we present insightful interpretations and properties of the Regularized Laplacian method.
We analyse important limiting cases in Section~5. Then, in Section~6 we discuss various numerical approaches
to compute the classification functions and show by numerical examples that the performance of the Regularized
Laplacian method is better or comparable with the leading semi-supervised methods. Section~7 concludes the
paper with directions for future research.

\section{Notations and method formulation}

Suppose one needs to classify $N$ data points (nodes) into $K$ classes and assume $P$ data points are labelled.
That is, we know the class to which each labelled point belongs.
Denote by $V_k$\x{-,} the set of labelled points in class $k=1,...,K$. Of course, $|V_1|+...+|V_K|=P$.

The graph-based semi-supervised learning approach uses a weighted graph $G=(V,A)$ connecting data
points, where $V$, $|V|=N$, denotes the set of nodes and $A$ denotes the weight (similarity) matrix.
In this work we assume that $A$ is symmetric and the underlying graph is connected.
Each element $a_{ij}$\x{For compactness and unification with Y_{ik}, w_{ij} removed comma between the subscripts i,j of matrix entries.} represents the degree of similarity between data points $i$ and~$j$.
Denote by $D$ the\x{-a} diagonal matrix with its \((i,i)\)-element equal to the sum
of the \(i\)-th row of matrix \(A\): \(d_{i}=\sum_{j=1}^{N}a_{ij}\).
We denote by $L=D-A\x{-W}$ the Standard (Combinatorial) Laplacian associated with the graph~$G$.

Define an \(N\times K\) matrix \(Y\) as
\begin{equation}
Y_{ik}=
   \begin{cases}
        1, & \text{if $i \in V_k$, i.e., point $i$ is labelled as a class $k$ point,}\\
        0, & \text{otherwise.}
   \end{cases} \nonumber
\end{equation}
We refer to each column $Y_{\ast k}$ of matrix $Y$ as a \emph{labeling function}.\x{\emph}
Also define an \(N\times K\) matrix \(F\) and call its columns $F_{\ast k}$
\emph{classification functions}.\x{\emph} The general idea of the graph-based semi-supervised learning
is to find classification functions so that on the one hand they are close
to the corresponding labeling function and on the other hand they change
smoothly over the graph associated with the similarity matrix.
This general idea can be expressed by means of the following particular
optimization problem:
\begin{equation}
\label{OptProb}
\min_{F}\left\{\x{left} \sum_{k=1}^K (F_{\ast k}-Y_{\ast k})^T (F_{\ast k}-Y_{\ast k})
+ \beta \sum_{k=1}^K F_{\ast k}^T L F_{\ast k} \right\}\x{\right},
\end{equation}
where $\beta \in (0,\infty)$ is a regularization parameter.
The regularization parameter $\beta$ represents a trade-off between
the closeness of the classification function to the labeling
function and its smoothness.

Since the Laplacian $L$ is positive-semidefinite and the second term in (\ref{OptProb})
is strictly convex, the optimization problem (\ref{OptProb}) has a unique solution
determined by the stationarity condition
$$
2 (F_{\ast k}-Y_{\ast k})^T + 2 \beta F_{\ast k}^T L = 0, \quad k=1,...,K,
$$
which gives
\begin{equation}
\label{eq:FRegLap}
F_{\ast k} = (I + \beta L)^{-1} Y_{\ast k}, \quad k=1,...,K.
\end{equation}
The matrix $Q_\beta = (I + \beta L)^{-1}$ is known as {\it Regularized Laplacian kernel\x{4}\/} of the graph \cite{KL02,SK03}
and can be related to the matrix forest theorems \cite{CS97,AgaChe01} and stochastic matrices \cite{AgaChe01}.\x{4}
The classification functions $F_{\ast k}, k=1,...,K,$ can be obtained either by numerical
linear algebra methods (e.g., power iterations) applied to (\ref{eq:FRegLap}) or by numerical
optimization methods applied to (\ref{OptProb}). We elaborate on numerical methods in Section~6.
Once the classification functions are obtained, the points are classified according to the rule
$$
F_{ik} > F_{ik'}, \forall k' \neq k \quad \Rightarrow \quad \mbox{Point $i$ is classified into class $k$.}
$$
The ties can be broken in arbitrary fashion.


\section{Related approaches}

Let us discuss a number of related approaches. First, we discuss formal relations and
in the numerical examples section we compare the approaches on some benchmark examples.

\subsection{Relation to heat kernels}

The authors of \cite{CY99,CY00} first introduced and studied the properties of the heat kernel
based on the normalized Laplacian. Specifically, they introduced the kernel
\begin{equation}
\label{eq:KerNormLap}
{\cal H}(t) = \exp (-t{\cal L}),
\end{equation}
where
$$
{\cal L} = D^{-1/2} L D^{-1/2}
$$
is the normalized Laplacian. Let us refer to ${\cal H}(t)$ as the \emph{normalized heat kernel}.\x{the, \emph}
Note that the normalized heat kernel can be obtained as a solution of the following
differential equation
$$
{\cal \dot{H}}(t)=-{\cal L}{\cal H}(t),
$$
with the initial condition ${\cal H}(0)=I$. Then, in \cite{C07} the PageRank heat kernel was introduced
\beq
\label{e_PRhk}
\Pi(t) = \exp(-t(I-P)),
\eeq
where
\beq
\label{e_Pst}
P=D^{-1}A,
\eeq
is the transition probability matrix of the \emph{standard random walk\/}
on the graph. In \cite{C09} the PageRank heat kernel was applied to local graph partitioning.

In \cite{KL02} the heat kernel based on the standard Laplacian
\begin{equation}
\label{HeatKer}
H(t) = \exp(-tL),
\end{equation}
with $L=D-A$, was proposed as a kernel in the support vector machine
learning method. Then, in \cite{ZGL03} the authors proposed a semi-supervised learning
method based on the solution of a heat diffusion equation with Dirichlet boundary conditions.
Equivalently, the method of \cite{ZGL03} can be viewed as the minimization of the second
term in (\ref{OptProb}) with the values of the classification functions $F_{\ast k}$\x{$...$} fixed on the labelled
points. Thus, the proposed approach (\ref{OptProb}) is more general as it can be viewed as a Lagrangian
relaxation of \cite{ZGL03}. The results of the method in \cite{ZGL03} can be retrieved with a particular
choice of the regularization parameter.

\newpage

\subsection{Relation to the generalized semi-supervised learning\\ method}

In \cite{AGMS12} the authors proposed a generalized optimization framework for graph
based semi-supervised learning methods
\begin{equation}\label{eq:genopt}
\min_{F} \left\{\sum_{i=1}^N \sum_{j = 1}^N w_{ij}\| {d_{i}}^{\sigma-1} F_{i \ast}  - {d_{j}}^{\sigma-1}F_{j \ast}\|^2
+ \mu\sum_{i=1}^N {d_{i}}^{2\sigma-1} \| F_{i \ast} - Y_{i \ast}\|^2 \right\},\x{\left\right}
\end{equation}
where $w_{ij}$ are the entries of a \emph{weight matrix} $W=(w_{ij})$ which is a function of $A$
(in particular, one can also take $W=A$).\x{}

In particular, with $\sigma=1$ we retrieve the transductive semi-supervised learning method \cite{ZB07},
with $\sigma=1/2$ we retrieve the semi-supervised learning with local and global consistency
\cite{Zetal04} and with $\sigma=0$ we retrieve the PageRank based method \cite{Aetal08}.

The classification functions of the generalized graph based semi-supervised learning are given
by
$$
F_{\ast k} = \frac{\mu}{2+\mu} \left(I - \frac{2}{2+\mu}D^{-\sigma} W D^{\sigma-1}\right)^{-1} Y_{\ast k},
\quad k=1,...,K.
$$
Now taking as the weight matrix $W = I - \tau L = I - \tau (D-A)$ (note that with this choice
of the weight matrix, the generalized degree matrix $D'=\diag(W\bm{1})$\x{} becomes the identity matrix), the above equation transforms to
$$
F_{\ast k} = \left(I + \frac{2\tau}{\mu} L \right)^{-1} Y_{\ast k},
\quad k=1,...,K,
$$
which is (\ref{eq:FRegLap}) with $\beta=2\tau/\mu$. It is very interesting to observe that
with the proposed choice of the weight matrix all the semi-supervised learning methods defined
by various $\sigma$'s coincide.

\section{Properties and interpretations of the Regularized Laplacian method}

There is a number of interesting interpretations and characterizations which we can provide for
the classification functions (\ref{eq:FRegLap}). These interpretations and characterizations will give
different insights about the Regularized Laplacian kernel\x{4} $Q_\beta$ and the classification functions (\ref{eq:FRegLap}).

\subsection{Discrete-time random walk interpretation}

The Regularized Laplacian kernel\x{4} $Q_\beta=(I+\beta L)^{-1}$ can be interpreted as the overall transition matrix of a random walk on the similarity graph $G$ with a geometrically distributed number of steps. Namely, consider a Markov chain whose states are our data points and the probabilities of transitions between distinct states are proportional to the corresponding entries
of the similarity matrix~$A$:
\beq
\label{graphMarkov}
\hat{p}_{ij}=\tau a_{ij}, \quad i,j=1\cdc N,\;\;i\ne j,
\eeq
where $\tau > 0$ is a sufficiently small parameter.
Then the diagonal elements of the transition matrix $\hat{P}=(\hat{p}_{ij})$ are
\beq
\label{dia_trans}
\hat{p}_{ii}=1-\sum_{j\ne i}\tau a_{ij},\quad i=1\cdc N
\eeq
or, in the matrix form,
\beq
\label{e_transM}
\hat{P} = I - \tau L.
\eeq

The matrix $\hat{P}$ determines a random walk on $G$ which differs from the ``standard'' one defined by~\eqref{e_Pst} and related to the PageRank heat kernel~\eqref{e_PRhk}.
As distinct from \eqref{e_Pst}, the transition matrix~\eqref{e_transM} is symmetric for every undirected graph; in general, it has a nonzero diagonal.
It is interesting to observe that $\hat{P}$ coincides with the weight matrix $W$ used for transformation of Subsection 3.2.

Consider a sequence of independent Bernoulli trials indexed by $0,1,2,\ldots$ with a certain success probability~$q$.
Assume that the number of steps, $K$, in a random walk is equal to the trial number of the first success.
And let $X_k$ be the state of the Markov chain at step $k$.
Then, $K$ is distributed geometrically:
\[
\Pr\{K=k\}=q(1-q)^k,\quad k=0,1,2,\ldots,
\]
and the transition matrix of the overall random walk after a random number of steps~$K$,
$Z=(z_{ij})$, $z_{ij}=\Pr\{X_K=j\mid X_0=i\},\quad i,j=1\cdc N,$ is given by
$$
Z = q \sum_{k=0}^\infty (1-q)^k \hat{P}^k = q \sum_{k=0}^\infty (1-q)^k (I-\tau L)^k
$$
$$
= q \left(I-(1-q)(I-\tau L)\right)^{-1} = \left(I + \tau (q^{-1}-1) L \right)^{-1}.
$$

\noindent
Thus, $Z=Q_\beta=(I + \beta L)^{-1}$ with $\beta=\tau(q^{-1}-1).$

This means that the $i$-th component of the classification
function can be interpreted as the probability of finding
the discrete-time random walk with transition matrix (\ref{e_transM}) in node $i$ after
the geometrically distributed number of steps with parameter $q$,
given the random walk started with the distribution $Y_{\ast k}/(\1^T\/Y_{\ast k})$.

\subsection{Continuous-time random walk interpretation}

Consider the differential equation
\begin{equation}
\label{eq:diffH}
\dot{H}(t) = -L H(t),
\end{equation}
with the initial condition $H(0)=I$. Also consider the standard continuous-time
random walk that spends exponentially distributed time in node $k$ with the expected
duration $1/d_k$ and after the exponentially distributed time moves to a new node $l$ with
probability $a_{kl}/d_k$. Then, the solution $h_{ij}(t)=\exp(-tL)$ of the differential
equation (\ref{eq:diffH}) can be interpreted as a probability to find the standard
continuous-time random walk in node $j$ given the random walk started from node $i$.
By taking the Laplace transform of (\ref{eq:diffH}) we obtain
\beq\label{e_Hs}\x{#}
H(s) = (sI + L)^{-1} = s^{-1}(I + s^{-1} L)^{-1}.\x{6m}
\eeq
Thus, the classification function (\ref{eq:FRegLap}) can be interpreted as the Laplace
transform divided by $1/s$, or equivalently the $i$-th component of the classification
function can be interpreted as a quantity proportional to the probability of finding
the random walk in node $i$ after exponentially distributed time with mean $\beta=1/s$
given the random walk started with the distribution $Y_{\ast k}/(\1^T\/Y_{\ast k})$.\x{maybe, more explicitly...}

\subsection{Proximity and distance properties}

As before, let $Q_\beta\!=\!(q_{ij}^\beta)_{N\!\times\! N}^{}$ be the Regularized Laplacian kernel\x{4}
$(I + \beta L)^{-1}$ of \eqref{eq:FRegLap}.

$Q_\beta$ determines a positive\x{4} \emph{$1$-proximity measure\/} \cite{CheSha98a} $s(i,j):=q^\beta_{ij},$ i.e., it satisfies \cite{CS97} the following conditions:\\
\indent $(1)$ for any $i\in V,$ $\sum_{k\in V}q^\beta_{ik}=1$ and\\ 
\indent $(2)$ for any $i,j,k\in V,$ $q^\beta_{ji}+q^\beta_{jk}-q^\beta_{ik}\le q^\beta_{jj}$ with a strict inequality whenever $i=k$ and $i\ne j$ (the \emph{triangle inequality for proximities}).

This implies \cite{CheSha98a} the following two important properties:
(a) $q^\beta_{ii}>q^\beta_{ij}$ for all $i,j\in V$ such that $i\ne j$ (\emph{egocentrism property});
(b) $\rho_{ij}^\beta:=\beta(q_{ii}^\beta+q_{jj}^\beta-q_{ij}^\beta-q_{ji}^\beta)$
is\footnote{Cf.\ the cosine law~\cite{Critchley88} and the inverse covariance mapping~\cite[Section\:5.2]{DezaLaurent97}.} a distance on~$V.$ Because of the forest interpretation of $Q_\beta$ (see Section~\ref{s_forest}), it is called the \emph{adjusted forest distance}. The distances $\rho_{ij}^\beta$ have a twofold connection with the \emph{resistance distance\/} $\tilde\rho_{ij}$ on~$G$
\cite{CheSha00}.
First, $\lim_{\beta\to\infty}\rho^\beta_{ij}=\tilde\rho_{ij},\;i,j\in V.$ 
Second, let $G^\beta$ be the weighted graph such that:\x{4:} $V(G^\beta)=V(G)\cup\{0\},$ the restriction of $G^\beta$ to $V(G)$ coincides with $G$, and $G^\beta$ additionally contains an edge $(i,0)$ of weight~$1/\beta$ for each node $i\in V(G)$. Then it follows that $\rho^\beta_{ij}(G)=\tilde\rho_{ij}(G^\beta),\;i,j\in V.$
In the electrical interpretation of $G$, the weight $1/\beta$ of the edges $(i,0)$ is treated as conductivity, i.e., the lines connecting each node to the ``hub'' $0$ have resistance~$\beta.$
An interested reader can find more properties of the proximity measures determined by $Q_\beta$ in~\cite{CS97}.

Furthermore, every $Q_\beta,$ $\beta>0$ determines a \emph{transitional measure\/} on $V,$ which means \cite{Che11AAM} that:
$q^\beta_{ij}\,q^\beta_{\!jk}\le q^\beta_{ik}\,q^\beta_{\!jj}$ for all $i, j,k\in V$ with\x{4with} $q^\beta_{ij}\,q^\beta_{\!jk}=q^\beta_{ik}\,q^\beta_{\!jj}$ if and only if every path in $G$ from $i$ to $k$ visits~$j.$

It follows that $d^\beta_{ij}:=-\ln\left(q^\beta_{ij}/\sqrt{q^\beta_{ii}q^\beta_{jj}}\right)$ provides a distance on~$V.$ This distance is \emph{cutpoint additive\/}, that is, $d^\beta_{ij}+d^\beta_{jk}=d^\beta_{ik}$ if and only if every path in $G$ from $i$ to $k$ visits~$j.$ In the asymptotics, $d^\beta_{ij}$ becomes proportional to the shortest path distance and the resistance distance as $\beta\to0$ and $\beta\to\infty,$ respectively.

\subsection{Matrix forest characterization}
\label{s_forest}

By the \emph{matrix forest theorem\/} \cite{CS97,AgaChe01},\x{4} each entry $q_{ij}^\beta$ of $Q_\beta$ is equal to the specific weight of the spanning rooted forests that \emph{connect node $i$ to node $j$\/}\x{6m} in the weighted graph $G$ whose combinatorial Laplacian is~$L.$

More specifically, $q_{ij}^\beta=\F^\beta_{i\dashv j}/\F^\beta,$ where $\F^\beta$ is the total $\beta$-weight of all spanning rooted forests of $G,$ $\F^\beta_{i\dashv j}$ being the total $\beta$-weight of such of them that have node $i$ in a tree rooted at~$j.$ Here, the \emph{$\beta$-weight of a forest\/} stands for the product of its edges weights, each multiplied by~$\beta.$

Let us mention a closely related interpretation of the Regularized Laplacian kernel\x{4} $Q_\beta$ in terms of information dissemination~\cite{Che08DAM}.
Suppose that an information unit (an idea) must be transmitted through~$G$.
A {\em plan\/} of information transmission is a spanning rooted forest $\Ff$ in $G$:
the information unit is initially injected into the roots of $\Ff$; after that it comes to the other nodes
along the edges of~$\Ff$. Suppose that a plan is chosen at random: the probability of every choice is proportional to the $\beta$-weight of the corresponding forest. Then by the matrix forest theorem, the probability that the information unit arrives at $i$ \emph{from root~$j$} equals $q_{ij}^\beta=\F^\beta_{i\dashv j}/\F^\beta$.
This interpretation is particularly helpful in the context of machine learning for social networks.

\subsection{Statistical characterization}

Consider the problem of attribute evaluation from paired comparisons.

Suppose that each data point (node) $i$ has a \emph{value parameter} $v_i,$ and a series of paired comparisons $r_{ij}$
between the points is performed.
Let the result of $i$ in a comparison with $j$ obey the Scheff\'e linear statistical model~\cite{Scheffe52}
\beq\label{e_PCm}
E(r_{ij})=v_i-v_j,
\eeq
where $E(\cdot)$ is the mathematical expectation. The matrix form of \eqref{e_PCm} applied to an experiment is
$$
E(\bm r)=X\bm v,
$$
where $\bm v=(v_1\cdc v_N)^T,$ and $\bm r$ is the vector of comparison results, $X$ being the \emph{incidence matrix\/} (\emph{design matrix}\/, in terms of statistics): if the $k$th element of $\bm r$ is a comparison result of $i$ confronted to $j,$ then, in accordance with \eqref{e_PCm}, $x_{ki}=1,$ $x_{kj}=-1,$ and $x_{kl}=0$ for $l\not\in\{i,\,j\}.$

Suppose that $X$ is known, $\bm r$ being a sample, and the problem is to estimate~$\bm v$ up to a shift \cite[Section\:4]{Che94}. Then
\beq\label{e_re}
\bm{\tilde v}(\lambda)=(\lambda I+X^TX)^{-1}X^T\bm r
\eeq
is the well-known \emph{ridge estimate} of $\bm v,$ where $\lambda>0$ is the \emph{ridge parameter}. Denoting $\beta=\lambda^{-1}$ and $X^TX=L$ (it is easily verified that $X^TX$ is a Laplacian matrix whose $(i,j)$-entry {with $j\ne i$ is minus}\x{#} the number of comparisons between $i$ and~$j$) one has
\beq\label{e_res}
\bm{\tilde v}(\lambda)=(I+\beta L)^{-1}\beta X^T\bm r,
\eeq
i.e., the solution is provided by the same transformation based on the Regularized Laplacian kernel\x{4} as in~\eqref{eq:FRegLap} (cf.\ also~\eqref{e_Hs})\x{#}. Here, the weight matrix $A$ of $G$ contains the numbers 
of comparisons between nodes; $\bm s=X^T\bm r$ is the vector of the sums of comparison results of the nodes: $s_i=\sum_{j}r_{ij}-\sum_{j}r_{ji},$ where $r_{ij}$ and $r_{ji}$ are taken from~$\bm r,$ which has one entry (either $r_{ij}$ or $r_{ji}$) for each comparison result.

Suppose now that value parameter $v_i$ (belonging to an interval centered at zero) is a \emph{positive or negative\/} intensity of some property, and thus, $v_i$ can be treated as a signed membership of data point $i$ in the corresponding \emph{class.}\x{4\emph} The pairwise comparisons $\bm r$ are performed with respect to this property.
Then $\beta X^T\bm r=\beta\bm s$\x{4\bm} 
is a kind of labeling function or a crude correlate of membership in the above class, whereas \eqref{e_res} provides a refined measure of membership which takes into account proximity. Along these lines, \eqref{e_res} can be considered as a procedure of semi-supervised learning.

A Bayesian version of the model \eqref{e_PCm} enables one to interpret and estimate the ridge parameter $\lambda=1/\beta.$ Namely, assume that:
\\(i) the parameters $v_1\cdc v_N$ chosen at random from the universal set are independent random variables with zero mean and variance $\sigma_1^2$ and
\\(ii) for any vector $\bm v$, the errors in \eqref{e_PCm} are independent and have zero mean, their unconditional variance being $\sigma_2^2.$

It can be shown \cite[Proposition~4.2]{Che94} that under these conditions, the best linear predictors for the parameters $\bm v$ are the ridge estimators \eqref{e_res} with $\beta=\sigma_1^2/\sigma_2^2.$

The \emph{best linear predictors} for $\bm v$ are the $\tilde v_i$'s that minimize $E(\tilde v_i-v_i)^2$ among all statistics of the form
$\tilde v_i=c_i+C_i^T\bm r$ satisfying $E(\tilde v_i-v_i)=0.$

The variances $\sigma_1^2$ and $\sigma_2^2$ can be estimated from the experiment. In fact, there are many approaches to choosing the ridge parameter, see, e.g., \cite{Dorugade14,MunizKibria09} and the references therein.

\section{Limiting cases}

Let us analyse the formula (\ref{eq:FRegLap}) in two limiting cases: $\beta \to 0$ and
$\beta \to \infty$. If $\beta \to 0$, we have
$$
F_{\ast k} = (I - \beta L) Y_{\ast k} + \mbox{o}(\beta).
$$
Thus, for very small values of $\beta$, the method resembles\x{"Resembles" is too vague.} the\x{} nearest neighbour method
with the weight matrix $W = I - \beta L$. If there are many points situated more than
one hop away from any labelled point, the method cannot produce good classification
with very small values of $\beta$. This will be illustrated by the numerical experiments in Section~6.

Now consider the other case $\beta \to \infty$. We shall employ the Blackwell series
expansion \cite{B62,P94} for the resolvent operator $(\lambda I+L)^{-1}$ with $\lambda=1/\beta$ \x{$\lambda$ and $L$ instead of $s$ and $D-A$}
\begin{eqnarray}
(I + \beta L)^{-1}
&=&\lambda(\lambda I + L)^{-1}
\nonumber\\\label{Blackwell}
&=&\lambda\left(\frac{1}{\lambda}\frac{1}{N} \1\1^T + H - \lambda H^2 + ...\right), \x{#is it new?}
\end{eqnarray}
where $H = (L+\frac{1}{N}\1\1^T)^{-1} - \frac{1}{N}\1\1^T$ is the generalized (group) inverse
of the Laplacian. Since the first term in (\ref{Blackwell}) gives the same value
for all classes if $\1^T Y_{\ast k} = \1^T Y_{\ast l}$,\x{#\ast l} $k \neq l$ (which is typically
the case), the classification will depend on the entries of the matrix~$H$ and finally, of the matrix $(L+\frac{1}{N}\1\1^T)^{-1}$.
Note that the matrix $(L+\alpha \1\1^T)^{-1}$, with a sufficiently small positive $\alpha$, determines a proximity measure called \emph{accessibility via dense forests}. Its properties are listed in \cite[Proposition~10]{CheSha98}.
An interpretation of $H$ in terms of spanning forests can be found in \cite[Theorem~3]{CheSha98};
see also~\cite{KirklandNeumannShader97}.

The accessibility via dense forests violates a natural \emph{monotonicity\/} condition, as distinct from $(I+\beta L)^{-1}$ with a finite~$\beta.$ Thus, a better performance of the regularized Laplacian proximity measure with finite\x{4} values of $\beta$ can be expected.

For the sake of comparison, let us analyse the limiting behaviour of the heat kernels. For instance, let us consider the Standard
Laplacian heat kernel (\ref{HeatKer}), since it is also based on the Standard Laplacian. In fact, it is immediate to see that the Standard
Laplacian heat kernel has the same asymptotic as the Regularized Laplacian kernel. Namely, if $t \to 0$,
$$
H(t) = \exp (-tL) = I - t L +\mbox{o}(t).
$$
Similar expressions hold for the other heat kernels. Thus, for small values of $t$, the semi-supervised learning methods based on
heat kernels should behave as the nearest neighbour method.

Next consider the Standard Laplacian heat kernel when $t \to \infty$. Recall that the Laplacian $L=D-A$ is a positive definite symmetric
matrix. Without the loss of generality, we can denote and rearrange the eigenvalues of the Laplacian as $0=\lambda_1\le \lambda_2 \le ...$ and
the corresponding eigenvectors as $u_1,...,u_n$. Note that $u_1 = \1$. Thus, we can write
$$
H(t) = u_1 u_1^T + \sum_{i=2}^N \exp(-\lambda_i t) u_i u_i^T.
$$
We can see that for large values of $t$ the first term in the above expression is non-informative as in the case of the Regularized Laplacian
method and we need to look for the second order term. However, in contrast to the Regularized Laplacian kernel, the second order term
$\exp(-\lambda_2 t) u_2 u_2^T$ is a rank-one term and cannot in principle give correct classification in the case of more than two classes.
The second term of the Regularized Laplacian kernel $H$ is not a rank-one matrix and as mentioned above can be interpreted in terms
of proximity measures.

\section{Numerical methods and examples}

Let us first discuss various approaches for the numerical computation of the classification functions (\ref{eq:FRegLap}).
Broadly speaking, the approaches can be divided into linear algebra methods and optimization methods. One of the basic
linear algebra methods is the power iteration method. Similarly to the power iteration method described in \cite{AMT15},
we can write
$$
F_{\ast k} = (I + \beta D - \beta A)^{-1} Y_{\ast k},
$$
$$
F_{\ast k} = (I - \beta (I+\beta D)^{-1}A)^{-1} (I+\beta D)^{-1}Y_{\ast k},
$$
$$
F_{\ast k} = (I - \beta (I+\beta D)^{-1}DD^{-1}A)^{-1} (I+\beta D)^{-1}Y_{\ast k}.
$$
Now denoting $B:=\beta (I+\beta D)^{-1}D$ and $C:=(I+\beta D)^{-1}$, we can propose the
following power iteration method to compute the classification functions
\begin{equation}
\label{eq:poweriter}
F_{\ast k}^{(s+1)} = B D^{-1} A F_{\ast k}^{(s)} + C Y_{\ast k}, \quad s=0,1,... \ ,
\end{equation}
with $F_{\ast k}^{(0)} = Y_{\ast k}$. Since $B$ is a diagonal matrix with the diagonal entries less
than one, the matrix $B D^{-1} A$ is substochastic with the spectral radius less than one and the
power iterations (\ref{eq:poweriter}) are convergent. However, for large values of $\beta$ and $d_i$,
the matrix $B D^{-1} A$ can be very close to stochastic and hence the convergence rate of the power
iterations can be very slow. Therefore, unless the value of $\beta$ is small, we recommend to use
the other methods from numerical linear algebra for the solution of linear systems with
symmetric matrices (recall that L is a symmetric positive semi-definite matrix in the case of undirected graphs).
In particular, we tried the Cholesky decomposition method and the conjugate gradient method. Both methods
appeared to be very efficient for the problems with tens of thousands of variables. Actually, the
conjugate gradient method can also be viewed as an optimization method for the respective convex
quadratic optimization problem such as (\ref{OptProb}) and (\ref{eq:genopt}). A very convenient property
of optimization formulations (\ref{OptProb}) and (\ref{eq:genopt}) is that the objective, and consequently,
the gradient, can be written in terms of a sum over the edges of the underlying graph. This allows a very
simple (and with some software packages even automatic) parallelization of the optimization methods based
on the gradient. For instance, we have used the parallel implementation of the gradient based methods
provided by the NVIDIA CUDA sparse matrix library (cuSPARSE) \cite{CUDA} and it showed excellent performance.

Let us now illustrate the Regularized Laplacian method and compare it with some other state of the art
semi-supervised learning methods on two datasets: Les Miselables and Wikipedia Mathematical Articles.

\begin{figure}
\begin{center}
\subfigure[RL Method vs PR method]{
\resizebox*{6.8cm}{!}{\includegraphics{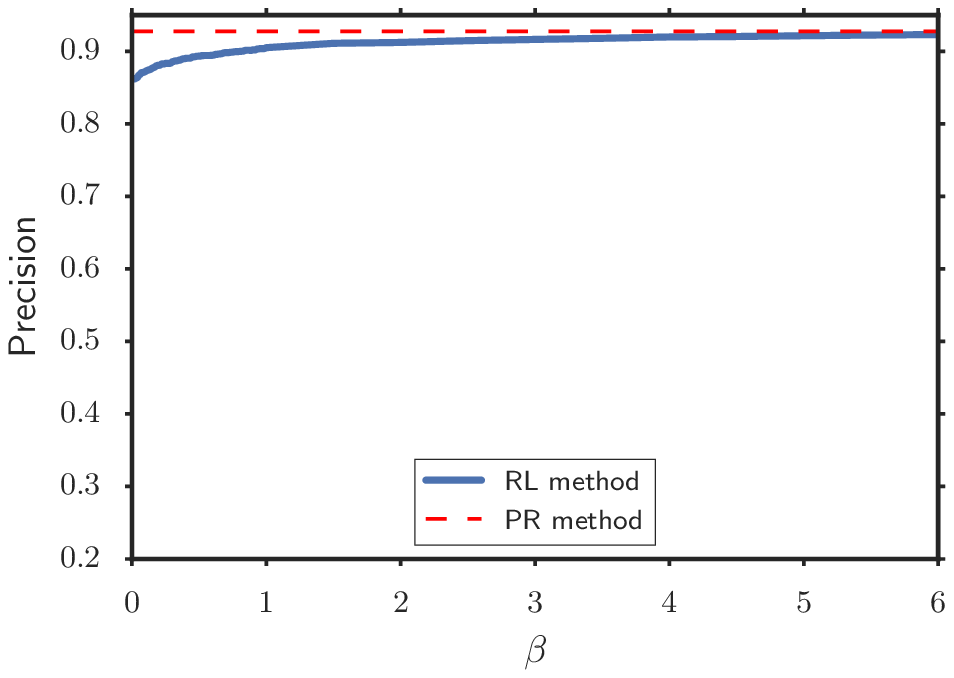}}}\hspace{5pt}
\subfigure[Heat Kernel methods vs PR method]{
\resizebox*{6.8cm}{!}{\includegraphics{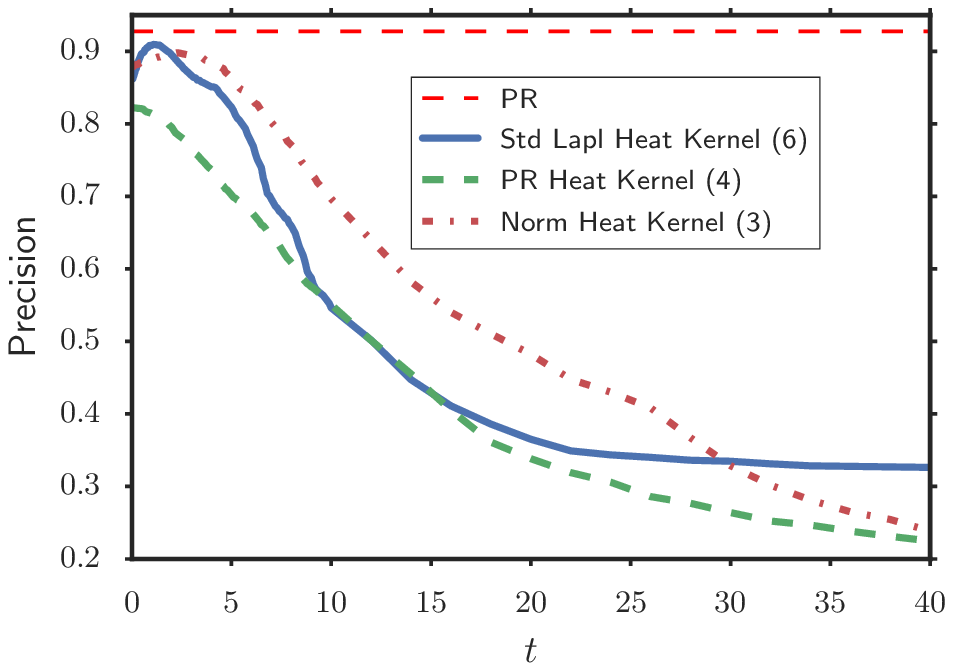}}}
\caption{Les Miserables Dataset. Labelled points are chosen randomly.} \label{fig:lesmisrand}
\end{center}
\end{figure}

The first dataset represents the network of interactions between major characters in the novel Les Miserables.
If two characters participate in one or more scenes, there is a link between these two characters. We consider
the links to be unweighted and undirected. The network of the interactions of Les Miserables characters
has been compiled by Knuth \cite{Knuth1993}. There are 77 nodes and 508 edges in the graph. Using
the betweenness based algorithm of Newman and Girvan \cite{Newman2004} we obtain 6 clusters which can be identified
with the main characters: Valjean (17), Myriel (10), Gavroche (18), Cosette (10), Thenardier (12), Fantine (10),
where in brackets we give the number of nodes in the respective cluster. First, we generate randomly (100 times)
labeled points (two labeled points per class). In Figure~\ref{fig:lesmisrand} we plot average precision as a
function of parameter $\beta$. In \cite{AGMS12,AGS13} it was observed that the PageRank based semi-supervised
method (obtained by taking $\sigma=0$ in (\ref{eq:genopt})) is the only method among a large family of semi-supervised
methods which is robust to the choice of the labelled data \cite{Aetal08,AGMS12,AGS13}. Thus, we compare the Regularized Laplacian method
with the PageRank based method. As we can see for Figure~\ref{fig:lesmisrand}.(a), the performance of the Regularized
Laplacian method is comparable to that of the PageRank based method on Les Miserables dataset.
The horizontal line in Figure~\ref{fig:lesmisrand}.(a) corresponds to the PageRank based method with the best choice
of the regularization parameter or the restart probability in the context of PageRank.
Since the Regularized Laplacian method is based on graph Laplacian, we also compare it in Figure~\ref{fig:lesmisrand}.(b)
with the three heat kernel methods derived from variations of the graph Laplacian. Specifically, we consider the three time-domain kernels based on various Laplacians: Standard Heat kernel (\ref{HeatKer}), Normalized Heat kernel (\ref{eq:KerNormLap}), and PageRank Heat kernel (\ref{e_PRhk}).
For instance, in the case of the Standard Heat kernel the classification functions are given by $F_{\ast k} = H(t) Y_{\ast k}$.
It turns out that all the three time-domain heat kernels are very sensitive to the value of the chosen time, $t$.
Even though there are parameter settings that give similar performances of Heat kernel methods and the Regularized Laplacian method,
the Regularized Laplacian method has a large plateau for values of $\beta$ where the good performance of the method is assured.
Thus, the Regularized Laplacian method is more robust with respect to the parameter setting than the heat kernel methods.

\begin{figure}
\begin{center}
\includegraphics[width=.5\textwidth]{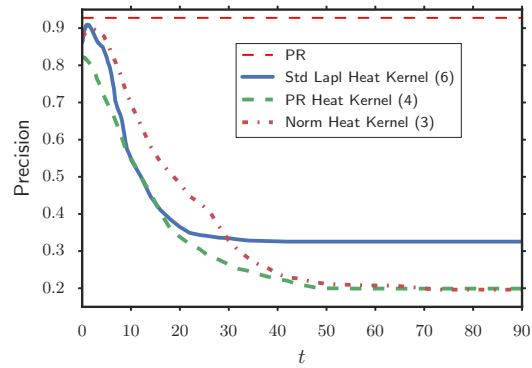}
\caption{Les Miserables Dataset. Heat Kernel methods vs PR method, larger $t$.}
\label{fig:largerts}
\end{center}
\end{figure}

To see better the behaviour of the heat kernel methods for large values of $t$, we have chosen a larger interval for $t$
in Figure~\ref{fig:largerts}. The performance of the heat kernel methods degrades quite significantly for large values of $t$.
This is actually predicted by the asymptotics given in Section~5. Since we have more than two classes, the heat kernels with
rank-one second order asymptotics are not able to distinguish among the classes. All heat kernel methods as well as the
Regularized Laplacian method show a deterioration in performance for small values of $t$ and $\beta$. This was predicted
in Section~5, as all the methods start to behave like the nearest neighbour method. In particular, as follows from the
asymptotics of Section~5 and can be observed in the figures the Standard Laplacian heat kernel method and the Regularized
Laplacian method shows exactly the same performance when $t \to 0$ and $\beta \to 0$.

\begin{figure}
\begin{center}
\subfigure[RL Method vs PR method]{
\resizebox*{6.8cm}{!}{\includegraphics{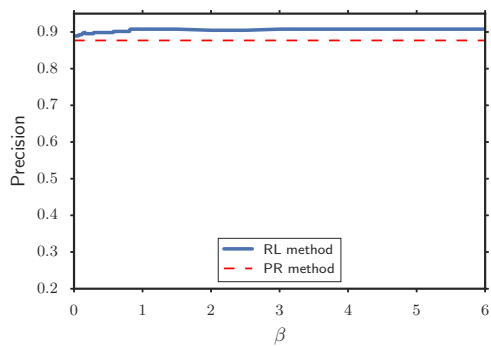}}}\hspace{5pt}
\subfigure[Heat Kernel methods vs PR method]{
\resizebox*{6.8cm}{!}{\includegraphics{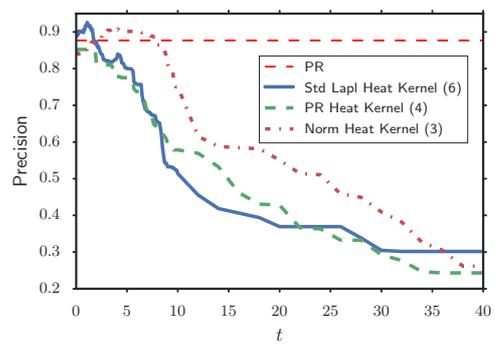}}}
\caption{Les Miserables Dataset. Labelled points are chosen with large degrees.} \label{fig:lesmisgs}
\end{center}
\end{figure}

It was observed in \cite{AGS13} that taking labelled data points with large (weighted) degree is typically
beneficial for the semi-supervised learning methods. Thus, we now label randomly two points out of three points
with maximal degree for each class. The average precision is given in Figure~\ref{fig:lesmisgs}.(a). We also test
heat kernel based methods with the same labelled points, see Figure~\ref{fig:lesmisgs}.(b). One can see that if
we choose the labelled points with large degree, the Regularized Laplacian Method outperforms the PageRank based
method. Some heat kernel based methods with large degree labelled points also outperform the PageRank based method
but their performance is much less stable with respect to the value of parameter $t$.

Next, we consider the second dataset consisting of Wikipedia mathematical articles. This dataset is derived from the English language
Wikipedia snapshot (dump) from January 30, 2010\footnote{\textbf{\texttt{http://download.wikimedia.org/enwiki/20100130}}}.
The similarity graph is constructed by a slight modification of the hyper-text graph. Each Wikipedia article typically contains links to other Wikipedia articles which are used to explain specific terms and concepts. Thus, Wikipedia forms a graph whose nodes represent articles and whose edges represent hyper-text inter-article links. The links to special pages (categories, portals, etc.) have been ignored. In the present experiment we did not use the information about the direction of links, so the similarity graph in our experiments is undirected. Then we have built a subgraph with mathematics related articles, a list of which was obtained from ``List of mathematics articles'' page from the same dump. In the present experiments we have chosen
the following three mathematical classes: ``Discrete mathematics'' (DM), ``Mathematical analysis'' (MA), ``Applied mathematics'' (AM). With the help of AMS MSC Classification\footnote{\textbf{\texttt{http://www.ams.org/mathscinet/msc/msc2010.html}}} and experts we have classified related Wikipedia mathematical articles into the three above mentioned classes. As a result, we obtained three imbalanced classes DM (106), MA (368) and AM (435). The subgraph induced by these three topics
is connected and contains 909 articles. Then, the similarity matrix $A$ is just the adjacency matrix of this subgraph.

\begin{figure}
\begin{center}
\subfigure[RL Method vs PR method]{
\resizebox*{6.8cm}{!}{\includegraphics{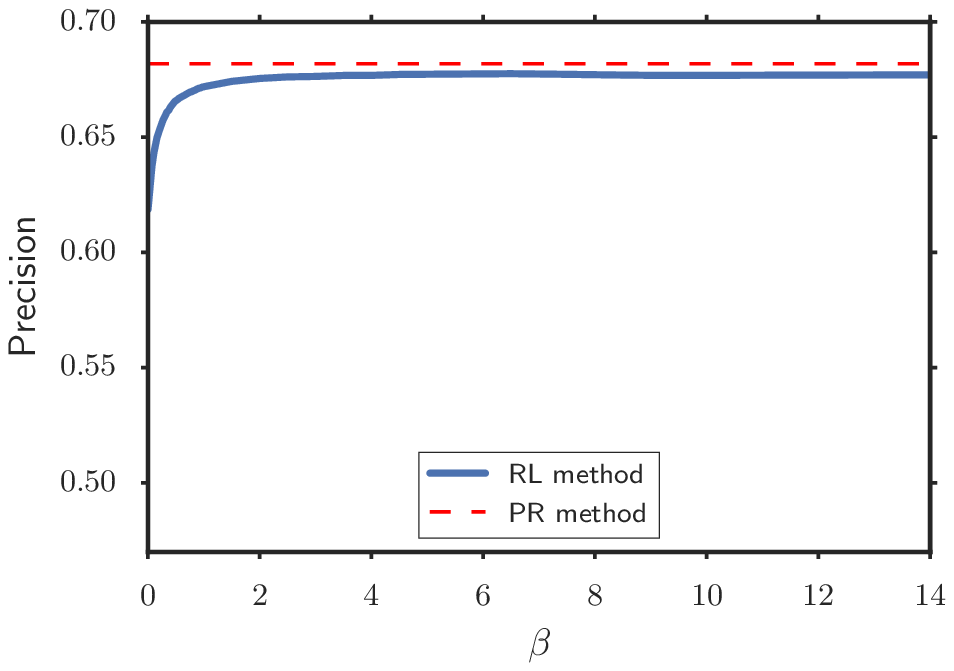}}}\hspace{5pt}
\subfigure[Heat Kernel methods vs PR method]{
\resizebox*{6.8cm}{!}{\includegraphics{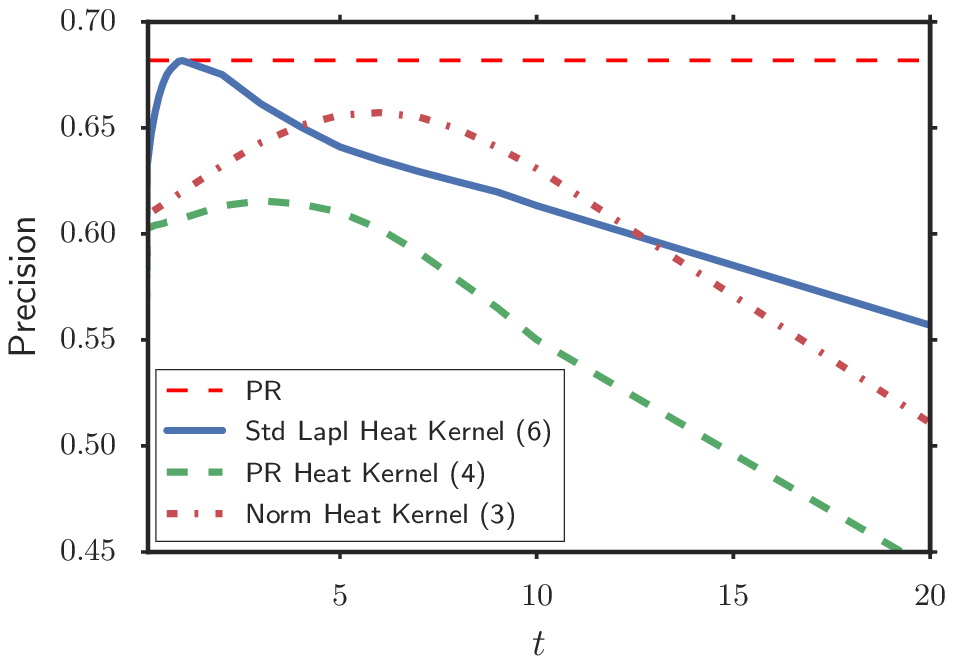}}}
\caption{Wiki Math Dataset. Labelled points are chosen randomly.} \label{fig:wikimathrand}
\end{center}
\end{figure}

First, we have chosen uniformly at random 100 times 5 labeled nodes for each class. The average precisions corresponding
to the Regularized Laplacian method and the PageRank based method are plotted in Figure~\ref{fig:wikimathrand}.(a).
We also provide the results for the three heat kernel based methods in Figure~\ref{fig:wikimathrand}.(b).
As one can see, the results of Wikipedia Mathematical articles dataset are consistent with the results of Les Miserables
dataset.

\begin{figure}
\begin{center}
\subfigure[RL Method vs PR method]{
\resizebox*{6.8cm}{!}{\includegraphics{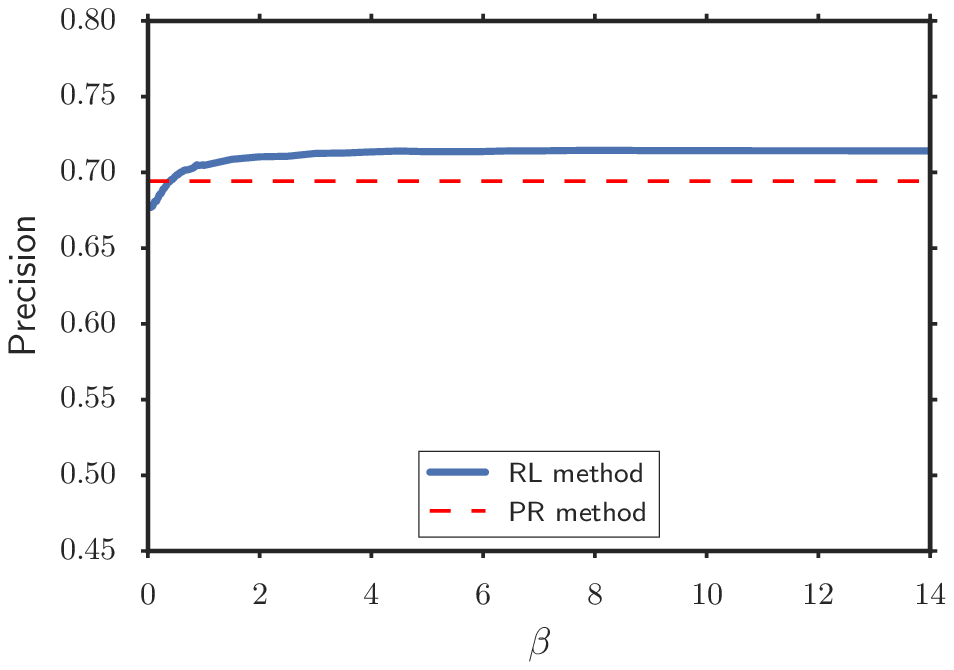}}}\hspace{5pt}
\subfigure[Heat Kernel methods vs PR method]{
\resizebox*{6.8cm}{!}{\includegraphics{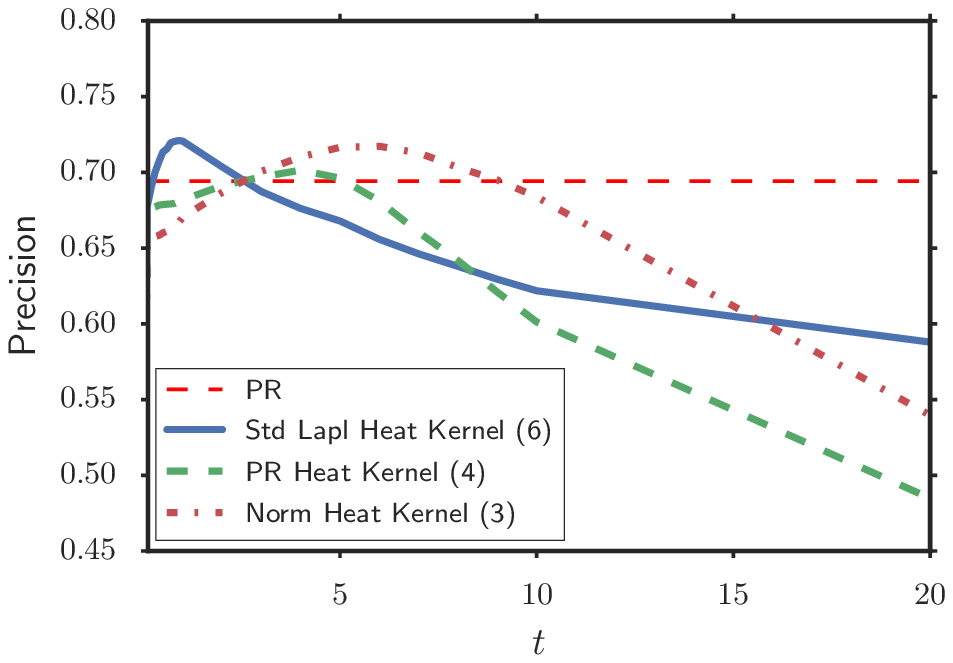}}}
\caption{Wiki Math Dataset. Labelled points are chosen with large degree.} \label{fig:wikimathgs}
\end{center}
\end{figure}

Then, for each class out of 10 data points with largest degrees we choose 5 points and average the results.
The average precisions for the Regularized Laplacian method, PageRank based method and for the three heat kernel
based methods are plotted in Figure~\ref{fig:wikimathgs}. The results are again consistent with the corresponding
results for Les Miserables dataset.
We would like to mention that for the computations in the Wiki Math dataset with many parameter settings and
extensive averaging using NVIDIA CUDA sparse matrix library (cuSPARSE) \cite{CUDA} were noticeably faster
than using numpy.linalg.solve calling LAPACK routine {\tt \_gesv}.

Finally, we would like to recall from Subsection~4.5 that a good value of $\beta$ can be provided by the
ratio $\sigma_1^2/\sigma_2^2$, where $\sigma_1^2$ is the variance related to the data points and $\sigma_2^2$
is the variance related to the paired comparison between points. We can argue that $\sigma_1^2$ is naturally
large and the paired comparisons between points can be performed with much more certainty, and hence, $\sigma_2^2$
is small. This gives a statistical explanation why it is good to take relatively large values for the
parameter $\beta$ in the Regularized Laplacian method.

\section{Conclusions}

We have studied in detail the semi-supervised learning method based on the Regularized Laplacian.
The method admits both linear algebraic and optimization formulations. The optimization formulation
appears to be particularly well suited for parallel implementation. We have provided various
interpretations and proximity-distance properties of the Regularized Laplacian graph kernel.
We have also shown that the method is related to the Scheff\'e linear statistical model.
The method was tested and compared with the other state of the art semi-supervised learning methods
on two datasets. The results from the two datasets are consistent. In particular, we can conclude that
the Regularized Laplacian method is comparable in performance with the PageRank based method and outperforms
the related heat kernel based methods in terms of robustness.

Several interesting research directions remain open for investigation. It will be interesting to compare
the Regularized Laplacian method with the other semi-supervised methods on a very large dataset. We are
currently working in this direction. We observe that there is a large plateau of $\beta$ values for which
the Regularized Laplacian method performs very well. It will be very useful to characterize this plateau
analytically. Also, it will be interesting to understand analytically why the Regularized Laplacian method
performs better when the labelled points with large degree are chosen.


\newpage
\tableofcontents

\end{document}